%% file: main.tex
\documentclass{article}
\usepackage{ijcai24}

\usepackage{times}
\usepackage{soul}
\usepackage{url}
\usepackage{tabularx}
\usepackage[hidelinks]{hyperref}
\usepackage[utf8]{inputenc}
\usepackage[small]{caption}
\usepackage{graphicx}
\usepackage{amsmath}
\usepackage{amsthm}
\usepackage{booktabs}
\usepackage{algorithm}
\usepackage[switch]{lineno}

\title{Collaborative Information Dissemination with Graph-based Multi-Agent Reinforcement Learning}

\newcommand{\BibTeX}{\rm B\kern-.05em{\sc i\kern-.025em b}\kern-.08em\TeX}

\usepackage{tabularx}
\usepackage{makecell}
\usepackage{algorithm}
\usepackage{algpseudocode}
\usepackage{algorithmicx}
\usepackage{nameref}
\usepackage{amsmath,amssymb,amsfonts}
\usepackage{graphicx}
\usepackage{textcomp}
\usepackage{multirow}
\usepackage{xcolor}
\usepackage[inkscapearea=page]{svg}
\svgpath{{images/}} 
\usepackage{microtype} 
\usepackage[english]{babel} 
\usepackage{subcaption}
\usepackage{siunitx}
\usepackage{gensymb}
\usepackage{booktabs}
\usepackage{tikz}
 
\usepackage{csquotes}
\usepackage{adjustbox}
\usepackage{tikz}
\graphicspath{ {images/} }
\usepackage{glossaries}
\glsdisablehyper
\loadglsentries[main]{acronyms}

\usepackage{times}
\usepackage{soul}
\usepackage{url}
\usepackage[hidelinks]{hyperref}
\usepackage[utf8]{inputenc}
\usepackage[font=small]{caption}
\usepackage{amsmath}
\usepackage{amsthm}
\usepackage{booktabs}
\usepackage{algorithm}
\usepackage[switch]{lineno}

\usepackage[textsize=small,textwidth=3cm]{todonotes}
\definecolor{kentuckyblue}{RGB}{0, 93, 170}

\author{%
Raffaele Galliera\textsuperscript{$\dag$,$\star$},
Kristen Brent Venable\textsuperscript{$\dag$,$\star$},
Matteo Bassani\textsuperscript{$\dag$},
Niranjan Suri\textsuperscript{$\dag$,$\star$,$\ddag$} \\ 
\normalsize{\normalfont{\textsuperscript{$\dag$}Institute for Human \& Machine Cognition\\
{\textsuperscript{$\star$}Department of Intelligent Systems \& Robotics}, The University of West Florida\\ 
Pensacola, FL, USA \\
\textsuperscript{$\ddag$}US Army Research Laboratory\\
Adelphi, MD, USA \\
\{rgalliera, bvenable, mbassani, nsuri\}@ihmc.org}}
}

\begin{document}

\maketitle

\begin{abstract}
Efficient information dissemination is crucial for supporting critical operations across domains like disaster response, autonomous vehicles, and sensor networks. This paper introduces a Multi-Agent Reinforcement Learning (MARL) approach as a significant step forward in achieving more decentralized, efficient, and collaborative information dissemination. We propose a Partially Observable Stochastic Game (POSG) formulation for information dissemination empowering each agent to decide on message forwarding independently, based on the observation of their one-hop neighborhood. This constitutes a significant paradigm shift from heuristics currently employed in real-world broadcast protocols. Our novel approach harnesses Graph Convolutional Reinforcement Learning and Graph Attention Networks (GATs) with dynamic attention to capture essential network features. We propose two approaches, L-DyAN and HL-DyAN, which differ in terms of the information exchanged among agents. Our experimental results show that our trained policies outperform existing methods, including the state-of-the-art heuristic, in terms of network coverage as well as communication overhead on dynamic networks of varying density and behavior.
\end{abstract}

\section{Introduction}
\input{sections/introduction}

\section{Background}
    \input{sections/background}

    \paragraph{Optimized Flooding in Broadcast Networks}
        \label{sec:problem_formulation}
        \input{sections/problem_formulation}

\section{Related Work}
    \label{sec:realated_work}
    \input{sections/related_work}

\section{Method}\label{sec:method}

In this section, we describe our novel \gls{marl} approach for optimizing information flooding in dynamic broadcast networks. 
We start by presenting a \gls{posg} formulation and then introduce our learning method and our two architectures, \gls{ldyan} and \gls{hldyan}, designed to achieve efficient dissemination while requiring different degrees of communication. 

    \subsection{MARL Formulation}
        \input{sections/MARL-formulation-flooding}

    \subsection{Learning Approach}\label{subsec:l-dgn}
        \input{sections/global_relation_kernel}

\section{Experiments}\label{sec:results}
In this section, we detail our experimental design to evaluate our two methods against \gls{dgn} and the \gls{mpr} heuristic employed in \gls{olsr}. Finally, we discuss our results.

    \input{sections/experimental_results}

\section{Conclusion and Future Work}
In this work, we captured the problem of information dissemination in dynamic broadcast networks in a novel \gls{posg} formulation and proposed two \gls{marl} methods to solve the task, namely \gls{ldyan} and \gls{hldyan}. Our experiments showed how these methods outperform in terms of coverage and message efficiency both \gls{dgn} and a popular heuristic employed in widely adopted broadcast protocols.

Our future research agenda includes investigating more structured group communication tasks, where, for example, coverage is desired only for a subset of nodes or nodes with higher priority. We will also study methods to enable more controlled trade-offs between coverage and forwarded messages, as well as their application in deployed protocols for physical computer networks.
Orthogonally, we will investigate the application of our approach to the dissemination of information in other domains, such as social networks and computational social choice.
\appendix

\newpage

\bibliographystyle{named}
\bibliography{bibliography}

\input{sections/Supplementary}

\end{document}

%% file: acronyms.tex
\newacronym{rl}{RL}{Reinfocement Learning}
\newacronym{mdp}{MDP}{Markov Decision Process}
\newacronym{sac}{SAC}{Soft Actor-Critic}
\newacronym{ml}{ML}{Machine Learning}
\newacronym{ai}{AI}{Artificial Intelligence}
\newacronym{rl3}{RL-Zoo3}{RL Baselines3 Zoo}
\newacronym{sgd}{SGD}{Stochastic Gradient Descent}
\newacronym{marl}{MARL}{Multi-Agent Reinforcement Learning}
\newacronym{dl}{DL}{Deep Learning}
\newacronym{nn}{NN}{Neural Network}
\newacronym{gnn}{GNN}{Graph Neural Network}
\newacronym{gat}{GAT}{Graph Attention Network}
\newacronym{diga}{DiGA-DQN}{Dueling Graph Attention Deep Q-Network}
\newacronym{olsr}{OLSR}{Optimized Link State Routing}
\newacronym{cnn}{CNN}{Convolutional Neural Network}
\newacronym{sg}{SG}{Stochastic Game}
\newacronym{ctde}{CTDE}{Centralized Training Decentralized Execution}
\newacronym{dtde}{DTDE}{Decentralized Training Decentralized Execution}
\newacronym{gcn}{GCN}{Graph Convolutional Network}
\newacronym{mpr}{MPR}{Multipoint Relay}
\newacronym{ilp}{ILP}{Integer Linear Program}
\newacronym{mcds}{MCDS}{Minimum Connected Dominating Set}
\newacronym{cds}{CDS}{Connected Dominating Set}
\newacronym{posg}{POSG}{Partially Observable Stochastic Game}
\newacronym{dpomdp}{Dec-POMDP}{Decentralized Partially Observable Markov Decision Process}
\newacronym{pomdp}{POMDP}{Partially Observable Markov Decision Process}
\newacronym{manet}{MANET}{Mobile Ad-hoc Network}
\newacronym{vanet}{VANET}{Vehicular Ad-hoc Network}
\newacronym{ddqn}{DDQN}{Double Deep Q-Network}
\newacronym{td}{TD}{Temporal Difference}
\newacronym{mlp}{MLP}{Multi Layer Perceptron}
\newacronym{ldyan}{L-DyAN}{Local Dynamic Attention Network}
\newacronym{hldyan}{HL-DyAN}{Hyperlocal Dynamic Attention Network}
\newacronym{dgn}{DGN}{Graph Convolutional Reinforcement Learning}
\newacronym{tc}{TC}{Topology Control}
\newacronym{smf}{SMF}{Simplified Multicast Forwarding}

%% file: sections/introduction.tex
Group communication, implemented in a broadcast or multicast fashion, finds a natural application in different networking scenarios, such as \glspl{vanet}~\cite{4300825,9051462}, with the necessity to disseminate information about the nodes participating, e.g. identity, status, or crucial events happening in the network.
These systems can be characterized by congestion-prone networks and/or different resource constraints, such that message dissemination becomes considerably expensive if not adequately managed. For this matter, message forwarding calls for scalable and distributed solutions able to minimize the total number of forwards, while achieving the expected coverage. Moreover, modern broadcast communication protocols often require careful adjustments of their parameters before achieving adequate forwarding policies, which would otherwise result in sub-optimal performance in terms of delivery ratio and latency~\cite{10017772}.

Recently, researchers have considered learning communication protocols~\cite{10.5555/3157096.3157336} with \gls{marl}~\cite{Busoniu2010}. At its core, \gls{marl} seeks to design systems where multiple agents learn to optimize their objective by interacting with the environment and the other entities involved. Such tasks can be competitive, cooperative, or a combination of both, depending on the scenario. As agents interact within a shared environment, they often find the need to exchange information to optimize their collective performance. This has led to the development of communication mechanisms that are learned rather than pre-defined, allowing agents to cooperate better utilizing their learned signaling system.

Nevertheless, learning to communicate with \gls{marl} comes with several challenges. In multi-agent systems, actions taken by one agent can significantly impact the rewards and state transitions of other agents, rendering the environment more complex and dynamic, and ensuring that agents develop a shared and consistent communication protocol, is an area of active research. Methods such as CommNet~\cite{10.5555/3157096.3157348} and BiCNet~\cite{peng2017multiagent}, focus on the communication of local encodings of agents' observations. These approaches allow agents to share a distilled version of their perspectives, enabling more informed collective decision-making. ATOC~\cite{10.5555/3327757.3327828} and TarMAC~\cite{pmlr-v97-das19a} have ventured into the realm of attention mechanisms. By leveraging attention, these methods dynamically determine which agents to communicate with and what information to share, leading to more efficient and context-aware exchanges. Yet another approach, as exemplified by \gls{dgn}~\cite{Jiang2020Graph}, harnesses the power of \glspl{gnn} and attention mechanisms to model the interactions, relations, and communications between agents.

However, to the best of our knowledge, no \gls{marl}-based method involving proactive communication and \glspl{gnn} has been proposed to address the unique challenges of optimizing the process of information dissemination within a broadcast dynamic network. In such a scenario, nodes need to cooperate to spread the information by forwarding it to their immediate neighbors, which might change over time, while relying on their limited observation of the entire graph. Furthermore, their collaboration and ability to accomplish dissemination are bound by the limitations of the underlying communication channels. This means that both the quantity of forwarding actions and the amount of information exchange needed for effective cooperation are constrained and should be minimized.

\paragraph{Contributions} In this work, we introduce {\bf a novel \gls{posg} for optimized information dissemination} in dynamic broadcast networks, forming the basis for our \gls{marl} framework.\footnote{\href{https://github.com/RaffaeleGalliera/melissa}{https://github.com/RaffaeleGalliera/melissa}.}

To this end, we design a \gls{marl} algorithm to encourage \textbf{cooperation within dynamic neighborhoods} where node connections are frequently changing. Furthermore, we design and test {\bf two distinct architectures}, namely \gls{ldyan} and \gls{hldyan}, which require different levels of communication leveraging \gls{gat} with dynamic attention~\cite{brody2022how} and Dueling Q-Networks~\cite{10.5555/3045390.3045601}.

Our \textbf{experimental study} demonstrates our methods' efficacy in achieving superior network coverage across dynamic graphs in different scenarios, outperforming \gls{dgn} and the established \gls{mpr}~\cite{rfc7188} heuristic. Moreover, our approach operates on one-hop observations and empowers nodes to take independent forwarding decisions, unlike \gls{mpr}.

By exploring the potential of learning-based approaches for addressing information dissemination in dynamic networks, our work underscores the versatility of \gls{marl} in present and future, real-world applications such as information dissemination in social networks~\cite{10.1145/2503792.2503797}, space networks~\cite{doi:10.34133/2021/9826517}, and vehicle-safety-related communication services~\cite{6092508}.

%% file: sections/background.tex
\paragraph{Reinforcement Learning (RL)}
    provides solutions for sequential decision-making problems formulated as \glspl{mdp}~\cite{suttonBarto,puterman1994}. 
    The \gls{pomdp} extends the \gls{mdp} framework to scenarios where agents have limited or partial observability of the underlying environment and make decisions based on  belief states, which are probability distributions over the true states. 
    To this end, several methods have been proposed such as Deep Recurrent Q-Learning~\cite{HausknechtS15}.
    
    \paragraph{Multi-Agent Reinforcement Learning} For multi-agent systems the RL paradigm extends to \gls{marl}~\cite{Busoniu2010}, where multiple entities, potentially learners and non-learners, interact with the environment. In this context the generalization of \glspl{pomdp} leads to \gls{dpomdp}, characterized by the tuple $\langle \mathcal{I}, \mathcal{S}, {\mathcal{A}^i_{i \in I}, \mathcal{P}, {\mathcal{R}}, \mathcal{O}^i_{i \in I}, \gamma} \rangle$. Here, $\mathcal{I}$ represents the set of agents, $\mathcal{S}$ denotes the state space, ${\mathcal{A}^i}_{i \in \mathcal{I}}$ stands for the action space for each agent, $\mathcal{P}$ is the joint probability distribution governing the environment dynamics given the current state and joint actions, ${\mathcal{R}}$ denotes the reward function, and $\mathcal{O}^i_{i \in \mathcal{I}}$ represents the set of observations for each agent. Such game-theoretic settings are used to model fully cooperative tasks where all agents have the same reward function and share a common reward.
    
    A more general model, adopted in this work, is the Partially Observable Stochastic Game (POSG), where each agent receives an individual reward ${\mathcal{R}^i}_{i \in \mathcal{I}}$, allowing the definition of fully competitive and mixed tasks such as zero-sum and general-sum games~\cite{marl-book}.
Several MARL algorithms have been presented in the literature, addressing different tasks (cooperative, competitive, or mixed) and pursuing different learning goals such as stability or adaptation \cite{Busoniu2010,AhmedBCCDFGGGMP22}. 

\paragraph{Graph Neural Networks}

\glspl{gnn}~\cite{gnn} process graph structures, 
and enable predictions at the graph, node, or edge level.
This is achieved by combining function approximators such as \glspl{nn} with a \textit{Message Passing} mechanism, where a node $x_i$ aggregates over the immediate neighbors' features and combines its features with the aggregated information. Repeating this operation $N$ times convolves information over nodes  $N$ hops away. 
\glspl{gnn} have shown remarkable success in several domains, such as recommendation systems, drug discovery, and social network analysis. Recent advancements have introduced novel \gls{gnn} architectures, such as \gls{gcn}~\cite{kipf2017semisupervised}, GraphSAGE~\cite{graphSage}, and \gls{gat}~\cite{veličković2018graph}, which have improved the performance on various tasks. In this paper, we use GATs with dynamic attention \cite{brody2022how} to capture relevant features of communication networks.

\paragraph{Graph Convolutional Reinforcement Learning}
In  \gls{dgn}~\cite{Jiang2020Graph}, the dynamics of multi-agent environments are represented as a graph, where each agent is a node with a set of neighbors determined by specific metrics. 
In this approach, a key role is played by Relation Kernels and their capability to merge features within an agent's receptive field, which is increased with the number of graph convolutional layers, all while capturing detailed interactions and relationships between agents. 
Building upon this foundation, during training, a batch of experiences $\mathcal{B}$ is sampled and the following  loss is minimized:
\begin{equation}\label{eq:dgn_loss}
L(\theta) = \frac{1}{|\mathcal{B}|}\sum_{\mathcal{B}} \frac{1}{N} \sum_{i=1}^{N} (y_i - Q (O_{i,\mathcal{C}} , a_i ; \theta))^2
\end{equation}

where $N$ is the number of agents, and $O_{i, \mathcal{C}}$ is the observation of agent $i$ with the respective adjacency matrix $\mathcal{C}$.  
We build on \gls{dgn} and design novel \gls{marl} architectures for optimizing information dissemination in dynamic networks.

%% file: sections/problem_formulation.tex
A dynamic broadcast network can be represented as a dynamic graph
$\mathcal{G}(t) = (\mathcal{V}, \mathcal{E}(t))$, 
where each node represents a (possibly) mobile node and an edge between two nodes at time $t$ represents the two corresponding nodes being within each other's broadcasting range at that time. 
Hence, for every node \(v \in \mathcal{V}\), the set of its neighbors at time $t$ is defined as $\mathcal{N}_v(t) = \{u \in \mathcal{V}| (v, u) \in \mathcal{E}(t)\}$.
 
A main objective of broadcast communications over connected networks is called Optimized Flooding~\cite{994521} and it is achieved when the information emitted from a given node $v \in \mathcal{V}$ reaches every other node $u \neq v $, thanks to forwarding actions of a set of nodes $\mathcal{D} \subseteq \mathcal{V}$.
While maximizing coverage it is also desirable to minimize redundant transmissions, which might impact resource utilization, such as bandwidth, power consumption, and latency. 
From a graph-theoretic point of view this can be achieved by identifying a specific subset of nodes, called a \gls{mcds}, that will be tasked with forwarding the information. This task requires the introduction of a centralized entity with complete knowledge of the network state and has been shown to be NP-complete~\cite{DBLP:books/fm/GareyJ79}.
A much more efficient and realistic approach is to approximate the MCDS in a distributed manner, relying only on local observations of the network made from each node's perspective.
Indeed, this is the approach taken by the \gls{mpr} selection heuristic and our \gls{marl} approach.

%% file: sections/related_work.tex

The \gls{mpr} selection algorithm is a technique developed to efficiently disseminate information in \glspl{manet} and wireless mesh networks. It achieves this by having each node designate certain one-hop neighbors to forward messages arriving from them, thereby reducing the overall transmission load and preventing excessive network broadcasting. This process involves nodes exchanging \enquote{HELLO messages} to identify and select their \gls{mpr} sets, ensuring effective network coverage with minimal redundancy.

In real-world networking protocols, \gls{mpr} plays an essential role. For instance, in the \gls{olsr} protocol~\cite{rfc7188}, \gls{mpr} is fundamental in managing the distribution of \gls{tc} messages. Similarly, in \gls{smf}~\cite{10.17487/RFC6621}, \gls{mpr} is employed primarily for the efficient forwarding of multicast packets.

In this work, we compare our approach with the \gls{mpr} selection algorithm, as outlined in the standard \gls{olsr} implementation~\cite{rfc7188}, leveraging this algorithm as a baseline for distributed message dissemination in dynamic graph structures.\footnote{The complete \gls{mpr} selection algorithm used by \gls{olsr} is shown in Algorithm~\ref{alg:mpr-selection}-Appendix~\ref{sec:mpr-pseudo}} However, we define a completely different approach that leverages \gls{marl} and, unlike \gls{mpr}, only requires an anonymized knowledge of the one-hop neighborhood and empowers each agent to independently decide their message forwarding policy.


Recent work has considered \gls{marl} approaches in the context of communication networks \cite{kaviani-admr,kaviani-cq,kaviani2023deepmpr}. The more closely related to the presented work is DeepMPR~\cite{kaviani2023deepmpr} which addresses the optimization of specific networking metrics in the context of multicast networks. While related, both the problem considered and the approach taken here are different. 
In particular, we focus on coverage and forwarded message minimization rather than metrics such as goodput, which apply only to such domains.
Moreover, in contrast to \cite{kaviani2023deepmpr} where the proposed method utilizes PPO \cite{schulman2017proximal} and policies trained with observation and action spaces tailored to specific graph scenarios, we design a more general approach, introducing a novel \gls{posg} with a dynamic number of participating agents and a scalable graph-based solution, which employs two novel approaches based on \gls{dgn} and \gls{gat} with dynamic attention for capturing essential local features and relations among agents. 


%% file: sections/MARL-formulation-flooding.tex
 We envision the dissemination process discretized into timesteps and episodes starting with a source node transmitting the information (a message) to its immediate (one-hop) neighbors. Each node in the graph corresponds to an agent observing its one-hop neighborhood and their features. At every timestep, nodes that have received the message will sense their neighborhood and decide whether to forward it to its current one-hop neighbors or stay silent. However, agents do not have any control or information 
 on who will be part of 
 their neighborhood
 at the next time-step. 
 Finally, the agents' objective is to disseminate the message emitted from the source node, i.e. maximize the network coverage, while minimizing the amount of forwarding actions, i.e. the messages, required. 

 An agent becomes a meaningful actor, receiving appropriate reinforcement signals
 only once it receives the message and for a limited number of steps.
 We capture this by implementing two different elements of our \gls{posg}. On the one hand, we distinguish \enquote{Graph Episodes} from \enquote{Agent Episodes}, allowing the agents to dynamically enter and leave the game independently. For this purpose, upon message reception, we limit the Agent's Episode to a fixed number of steps (\textit{local horizon}) during which it decides whether to forward the message to its immediate neighbors or not. Graph Episodes, model the overall dissemination process, and terminate once every agent that has received the message has exhausted its local horizon. 
 Given the agents' asynchronous presence, reward signals are issued individually to each agent, but capture the necessary degree of cooperation within their neighborhoods.

In our formulation agents are anonymous (i.e. not identified by any ID) and sense 
only their immediate neighborhood, accessing the degree of connectivity of such neighbors and observing their forwarding behavior. This is far more parsimonious than what is required by \gls{mpr} that requires agents to obtain a complete, identified, two-hop knowledge. 


More specifically, given the broadcast network represented by graph \(\mathcal{G}_0 = (\mathcal{V}, \mathcal{E}_0)\) at time $t_0$, and node $n_s \in \mathcal{V}$, we define the \gls{posg} associated to the  optimized flooding of $\mathcal{G}_0$ with source $n_s$ and network update function $\mathcal{U}$, with the tuple  $\langle \mathcal{I}, \mathcal{S}, \mathcal{A}^i_{i \in I}, \mathcal{U}, \mathcal{P}, \mathcal{R}^i_{i \in I}, \mathcal{O}^i_{i \in I}, \gamma \rangle$, where:

\paragraph{Agents set ~$\mathcal{I}$.} 
Set $\mathcal{I}$ contains one agent for each node in $\mathcal{V}$. $\mathcal{I}$ is divided into three disjoint sets which are updated at every timestep $t$: the active set $\mathcal{I}_a(t)$, the done set $\mathcal{I}_d(t)$, and the idle set $\mathcal{I}_i(t)$. Agents in $\mathcal{I}_i(t)$ are inactive because they have not received the message yet.
At the beginning of the process, 
$\mathcal{I}_i(t)$ will contain all agents except the one associated with $n_s$.
Agents in $\mathcal{I}_d(t)$ are also inactive, after participating in the game and  
 terminating their Agent Episode once the local horizon is reached. $\mathcal{I}_a(t)$, instead, includes the set of agents actively participating in the game at timestep $t$. Agents in $\mathcal{I}_i(t)$ are moved to $\mathcal{I}_a$ at time step $t+1$, if they have been forwarded the information, hence starting their Agent Episode.

\paragraph{Actions $\mathcal{A}^i_{i \in I}$.} For any time step $t$, if agent $i$ is in $\mathcal{I}_a(t)$, then, $\mathcal{A}^i$ contains two possible actions: forward the information to their neighbors or stay silent. The action set for agents in 
$\mathcal{I}_i(t)$ and $\mathcal{I}_d(t)$ 
is, instead,  empty.

\paragraph{Environment Dynamics $\mathcal{P}$ and Network Update $\mathcal{U}$.}
The environment dynamics are defined by the transition function $\mathcal{P}: \mathcal{S} \times \mathcal{A}^1 \times \cdots \times \mathcal{A}^{|I|} \rightarrow \Delta(\mathcal{S})$, where $\Delta(\mathcal{S})$ represents the set of probability distributions over the state space $\mathcal{S}$.
In our POSG model, we incorporate a general stochastic network update function, $\mathcal{U}$, controlling how the edges of the network change over time. This element allows us to capture various dynamics such as agent mobility or other factors that may affect a network's connectivity.
More formally, at every timestep $t$, the graph structure is updated such that $\mathcal{G}_{t+1} = \mathcal{G}(\mathcal{V}, \mathcal{U}(\mathcal{E}_t) )$.

The message-forwarding mechanism is purposefully modeled as deterministic and, at each timestep $t$, if an active agent $i$ forwards the message, all nodes in $\mathcal{N}_i(t)$ will receive it.
\begin{figure}[ht]
  \centering
  \begin{minipage}[t]{0.3\linewidth} 
    \centering
    \includegraphics[width=\linewidth]{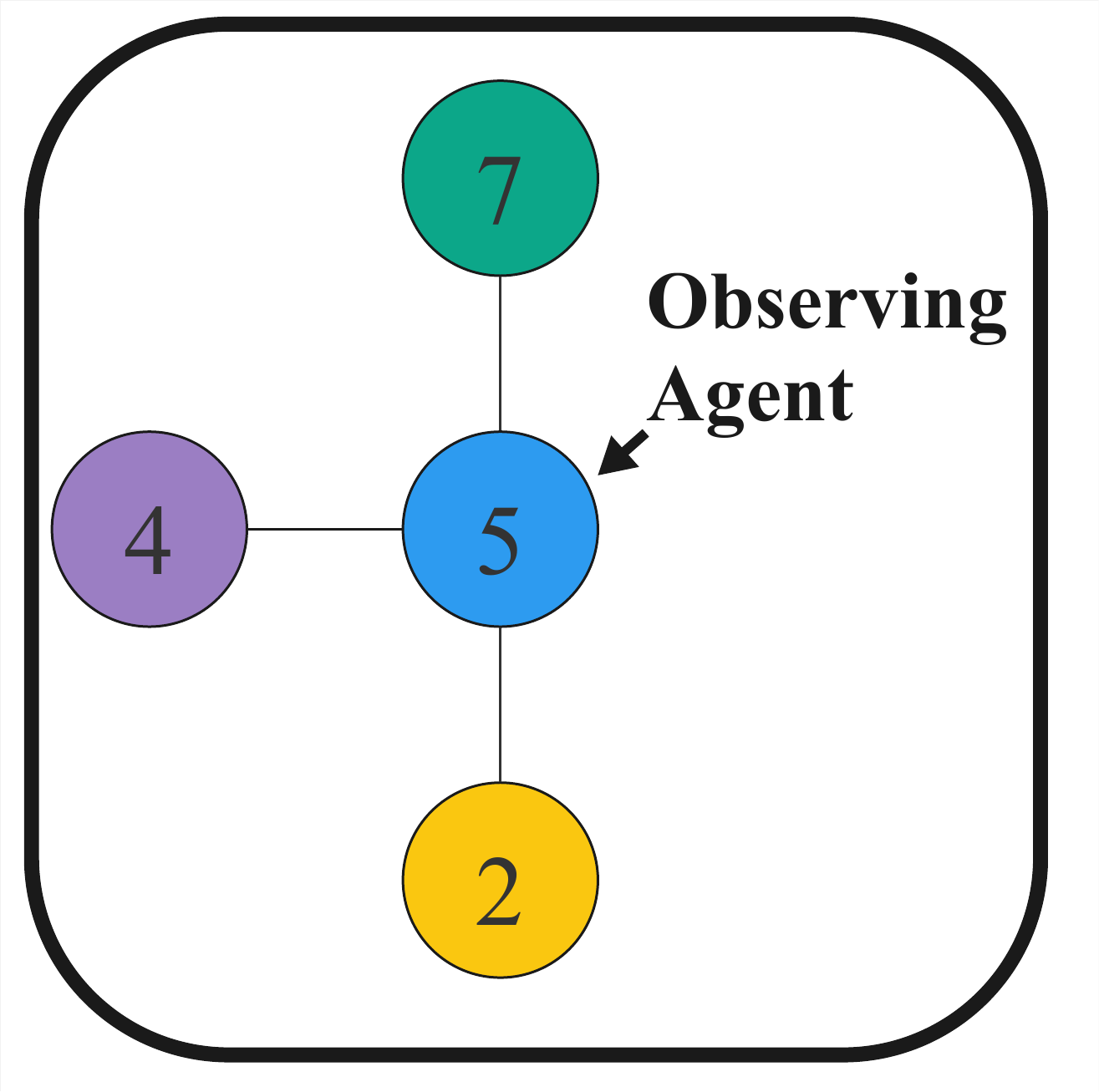} 
    \label{fig:image}
  \end{minipage}
  \hfill 
  \begin{minipage}[b]{0.68\linewidth} 
    \centering
    \input{tables/state}
  \end{minipage}
  \captionof{table}{Example of agent observation along with nodes features.
  }
  \label{table:features}

\end{figure}

\paragraph{Observations $\mathcal{O}^i_{i \in I}$ and State set $\mathcal{S}$.} 
Each node in the graph has a set of three features observable by other neighboring agents at each time step $t$: neighborhood size, the number of messages transmitted, and its last action. The agents' observations are represented as the graph describing their one-hop neighborhood and the features associated with each node in this local structure. As an example, Table \ref{table:features} depicts the neighborhood graph and observation 
of agent 5.
In our setting, a state $\mathcal{S}_t$ corresponds to the current graph structure $\mathcal{G}_t$ and the following information for each node: the features as shown in Table~\ref{table:features}, the set to which the agent belongs $\mathcal{I}_a$, $\mathcal{I}_d$, or $\mathcal{I}_i$, and the remaining steps of the local horizon for those in $\mathcal{I}_a$. 

\paragraph{Rewards ${\mathcal{R}^i}_{i \in I}$.} At the end of each step every agent in $\mathcal{I}_a$
is issued with a reward signal with positive and negative components. 
The positive term rewards the agent based on its two-hop coverage, i.e. how many one- and two-hop
neighbors have received the information.
One of two penalties might be issued, based on the agent's behavior. If the agent has forwarded during its last action, it will participate in a shared transmission cost, punishing the agent for the number of messages sent by its neighborhood. Otherwise, it will receive penalties based on the unexploited coverage potential of neighbors who have not yet received the information.
Formally, the reward signal for agent $i$ at time $t$ be defined as follows:
\begin{small}

\begin{equation}\label{eq:pos_rews}
r_{i,t} = \frac{\upsilon(\mathcal{M}_i(t), t)}{|\mathcal{M}_i(t)|} - p(i, t), \mathcal{M}_i(t) = \bigcup_{u \in \mathcal{N}_i(t) \cup \{i\}} \mathcal{N}_u(t) \setminus \{i\}
\end{equation}

\begin{equation}
    \label{eq:penalties}
    p(i, t) = \begin{cases} 
    
    m(\mathcal{N}_i(t), t), & \mbox{if } i \in \mathcal{T}(t) 
    
    \\ \mu(\mathcal{N}_i(t), t), & \mbox{if } i \in \mathcal{I}_a(t) \setminus \mathcal{T}(t)
    \end{cases} 
\end{equation}
\end{small}

In Equation~\ref{eq:pos_rews}, $\mathcal{M}_i(t)$ represents the set of two-hop neighbors of agent $i$ at $t$. $\upsilon(\mathcal{M}_i, t)$ denotes the number of them that by timestep $t$ have already received the message, while $p(i, t)$, defines the penalties assigned to agent $i$. The latter is further described in Equation~\ref{eq:penalties}, where $\mathcal{T}(t)$ is the set of active agents that have forwarded the message at least once. Here $m(\mathcal{N}_i(t), t)$ denotes the sum of the number of messages transmitted by the current neighborhood of agent $i$ by timestep $t$. The term $\mu(\mathcal{N}_i(t), t)$ instead defines the Maximum Normalized Coverage Potential of node $i$, which we define as:

\begin{small}

\begin{equation}
\mu(\mathcal{N}_i(t), t) = \frac{\max(\mathcal{C}_i(t))}{\sum \mathcal{C}_i(t)}
\end{equation}

\end{small}

\begin{small}
    \begin{equation}
        \mathcal{C}_i(t) = \{|\mathcal{N}_j(t)| : j \in \mathcal{N}_i(t) \cap \mathcal{I}_i(t)\}
    \end{equation}
\end{small}

On the one hand, we note that by assessing the ability of an agent's neighborhood to reach nodes beyond its immediate neighbors, Equation \ref{eq:pos_rews}, encourages agents to collectively cover more nodes through coordination within their vicinity.
On the other hand, the neighborhood-shared transmission steers the agents away from redundancy, promoting efficient dissemination. 
Finally, the Maximum Normalized Coverage Potential counterbalances the shared transmission costs, by hastening transmission to nodes with highly populated neighborhoods that have not yet been reached.

%% file: tables/state.tex
\begin{small}
\begin{tabularx}{\columnwidth}{ c | X | X | c }
\toprule
    Node & Number of Neighbors & Messages Sent & $A_t$ \\
\midrule
    \textbf{2} & 3 & 3 & 0 \\
    \textbf{4} & 1 & 1 & 1 \\
    \textbf{5} & 3 & 3 & 0 \\
    \textbf{7} & 4 & 0 & 0 \\
\bottomrule
\end{tabularx}
\end{small}

%% file: sections/global_relation_kernel.tex
\begin{figure*}[t]
\centering    
\includegraphics[width=\textwidth]{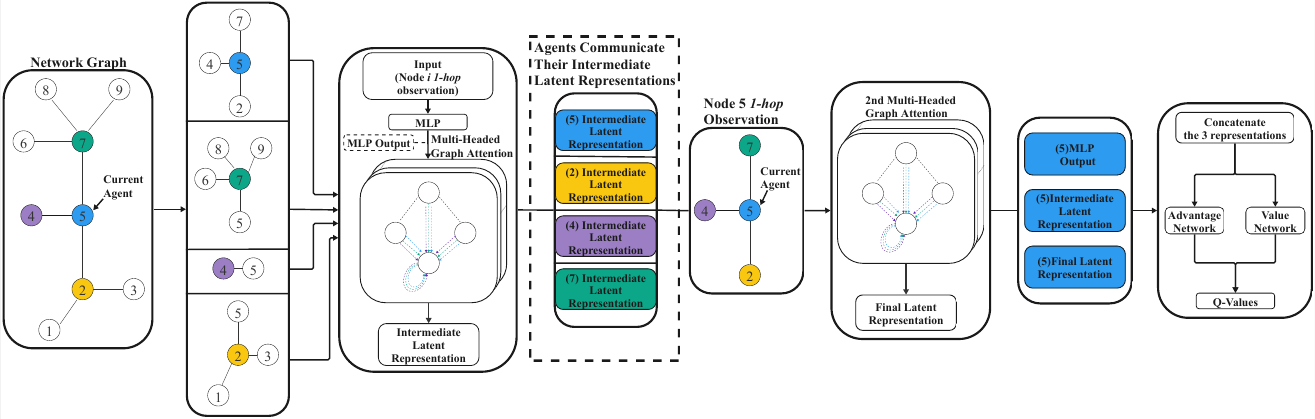}
    \caption{Information flow from a single agent observation (5) to the produced Q-Values in the \gls{ldyan} architecture.}
    \label{fig:learning_architecture}
\end{figure*}

The idea behind \gls{ldyan} and \gls{hldyan} is to encourage cooperation within dynamic neighborhoods, where links between nodes can form and/or disappear over time.
We therefore propose a loss function comprising neighborhood experiences, the usage of \gls{gat} layer(s) with dynamic attention~\cite{brody2022how}, and the presence of a dueling network to separately estimate the state-value and the advantages for each action~\cite{10.5555/3045390.3045601}. The choice of a \gls{gat} layer with dynamic attention is driven by its capability of capturing expressive attention mechanisms within a graph, a feature shown to be weaker in dot-product attention, as used in \gls{dgn}~\cite{brody2022how}. 



\subsubsection{Cooperative Dynamic Neighborhoods}\label{sec:local_horizon_estimation}
During training, at each timestep $t$, the tuple $(\mathcal{O}_{\mathcal{I}_a(t)}, \mathcal{A}_{\mathcal{I}_a(t)}, \mathcal{R}_{\mathcal{I}_a(t)}, \mathcal{O'}_{\mathcal{I}_a(t)})$ is stored in a circular replay buffer with a fixed length. $\mathcal{O}_{\mathcal{I}_a(t)}$ indicates the set of observations of all agents in $\mathcal{I}_a(t)$, $\mathcal{A}_{\mathcal{I}_a(t)}$ the set of actions taken by these agents, $\mathcal{R}_{\mathcal{I}_a(t)}$ is the set of rewards, and $\mathcal{O'}_{\mathcal{I}_a(t)}$ the set of observations of agents in $\mathcal{I}_a(t)$ at the next timestep.

At each training step, we sample a random batch $\mathcal{B}$ from the replay buffer, with every sample containing the experience of some agent $i$ and the ones of its current and active neighbors $\mathcal{N}_{i, \mathcal{I}_a(t)} = \mathcal{N}_i(t) \cap \mathcal{I}_a(t)$. The loss for each sample is computed not only based on the agent's own experience but also considering the experiences of its active neighbors. We denote $\mathcal{N}^{+i}_{i, \mathcal{I}_a(t)}= \mathcal{N}_{i, \mathcal{I}_a(t)} \cup \{i\}$ and define the loss function:

\begin{equation}
\mathcal{L}(\theta) = \frac{1}{|\mathcal{B}|}\sum_{\mathcal{B}} \frac{1}{|\mathcal{N}^{+i}_{i, \mathcal{I}_a(t)}|}
\sum_{j \in \mathcal{N}^{+i}_{i, \mathcal{I}_a(t)}}\left(y_t^j - Q(o^j, a^j;\theta)\right)^2,
\end{equation}

where, for each agent $j$,  $y_t^j$ is the target return and $Q(o^j, a^j;\theta)$ the predicted $Q$ value, parameterized with $\theta$, given the observation $o^j$ and action $a^j$. From this point onward, we will drop the superscript $j$ when referring to $o$, $a$, $r$, and $y$ as they will refer to a single experience.

Additionally, we take advantage of the agents' short-lived experiences and perform $n$-step returns, with $n$ equal to the local horizon ($k$). We note that the replay buffer is temporally sorted and organized such that every individual episode, ongoing or terminated with a length up to $k$, can be uniquely identified. If the buffer contains the remaining steps until the termination of the agent's episode, the $n$-step computation serves an unbiased value of the return:
$y_t = \sum_{i=0}^{k-t}\gamma^i r_{t+i}$.

If the trajectory stored in the buffer contains only the next $j$ steps  before termination, $y_t$ will be estimated as:
\begin{equation}
y_t = \sum_{i=0}^{j-1}\gamma^i r_{t+i} + \gamma^j Q(o_{t+i}, \text{argmax}_{a' \in \mathcal{A}} Q(o_{t+i}, a';\theta); \bar{\theta}),
\end{equation}

where $\theta$ is the current network and $ \bar{\theta}$ is the target network.

\subsubsection{Local-DyAN}

The first architecture we propose is depicted in Figure \ref{fig:learning_architecture} and consists of an encoder module comprised of three different stages: one \gls{mlp} followed by two multi-headed GATs~\cite{veličković2018graph} with dynamic attention~\cite{brody2022how}. The final latent representation will comprise the concatenation of each stage output, which is then fed to a dueling network decoding the final representation into the predicted $Q$ values. After each encoding stage, a $\mathrm{ReLU}$ activation function is applied. 

We now describe the flow from the agent's observation to the $Q$ values prediction and we show how it can be integrated into broadcast communication protocols.
Agent $i$'s observation at time $t$ is first fed to the \gls{mlp} encoding stage. This results in a learned representation of the features belonging to agent $i$ and its neighbors, denoted respectively $\mathbf{x}_i$ and $\mathbf{x}_j, \forall j \in \mathcal{N}_i(t)$. Following such encoding stage, the output of each of the $M$ attention heads of the first \gls{gat} is:

\begin{equation}
\label{eq:gat}
\mathbf{x}^{m}_i = \alpha^m_{i,i}\mathbf{W}\mathbf{x}_{i} +
\sum_{j \in \mathcal{N}_i(t)}\alpha^m_{i,j}\mathbf{W}\mathbf{x}_{j} 
\: \forall m \in \{0,...,M-1\},
\end{equation}

where the dynamic attention $\alpha^m$ for the tuple $(i, j)$, denoted as $\alpha^m_{i,j}$, is computed by: 

\begin{equation}
\label{eq:attention}
\alpha^m_{i,j} =
\frac{
\exp\left(\mathbf{a}^{\top}\mathrm{LeakyReLU}\left(\mathbf{W}
[\mathbf{x}_i \, \Vert \, \mathbf{x}_j]
\right)\right)}
{\sum_{k \in \mathcal{N}_i(t) \cup \{ i \}}
\exp\left(\mathbf{a}^{\top}\mathrm{LeakyReLU}\left(\mathbf{W}
[\mathbf{x}_i \, \Vert \, \mathbf{x}_k]
\right)\right)},
\end{equation}

where $\mathbf{a}$ and $\mathbf{W}$ are learned. We denote $\mathbf{\hat{X}}_i = \mathbf{x}^0_i||\mathbf{x}^1_i||...||\mathbf{x}^{M-1}_i$, where $||$ is the concatenation operator, as the concatenation of every attention output. Through message passing, each agent $i$ receives $\mathbf{\hat{X}}_j, \forall j \in \mathcal{N}_i(t)$.
These new representations
are fed to the second \gls{gat} layer, where the  computation follows the same logic seen in Equation~\ref{eq:gat} and~\ref{eq:attention}, producing the embedding $\mathbf{\hat{Z}}_i$.

Finally, the output of each encoding stage is concatenated in a final latent representation $\mathbf{H}_i$:

\begin{equation}
\mathbf{H}_i = \mathbf{x}_i||\mathbf{\hat{X}}_i||\mathbf{\hat{Z}}_i.
\end{equation}

At this point, $\mathbf{H}_i$ is fed to the two separate streams of the dueling network, namely the value network $V$ and the advantage network $A$, parameterized by two separate \glspl{mlp} with parameters $\alpha$ and $\beta$, respectively. Let us denote the parameterization previous to the dueling network, which produced the final latent representation $\mathbf{H}_i$ given $o$, as $\delta$. The predicted $Q$ values are then obtained as:

\begin{equation}\label{eq:duel}
\begin{split}
    Q(o, a; \delta, \alpha, \beta) &= V(o; \delta, \alpha) + 
    \\&+\left( A(o, a;\delta, \beta) - \frac{1}{|\mathcal{A}|} \sum_{a' \in \mathcal{A}} A(o, a';\delta, \beta) \right).    
\end{split}
\end{equation}

We note that the encoding process described above harmoniously integrates with the communication mechanisms present in protocols deployed in the real world, such as \gls{olsr}. We envision every node (agent) in the network, feeding its neighborhood structure through the encoding process described above. Subsequently, every agent shares their intermediate representation $\mathbf{\hat{X}}_i$ with their one-hop neighbors, in a similar way to how nodes communicate their \gls{mpr} sets in \gls{olsr}. Once the representations are collected, agents feed them to the second \gls{gat} layer, obtaining $\mathbf{H}_i$.
    
However, embedding the communication process generates a communication overhead of size proportional to $\mathbf{\hat{X}}_i$, an aspect which might need to be further minimized in bandwidth-constrained networks~\cite{10253585,10356270}. 
This observation leads us to our second approach.


\begin{figure}[t]
\centering    
\includegraphics[width=\columnwidth]{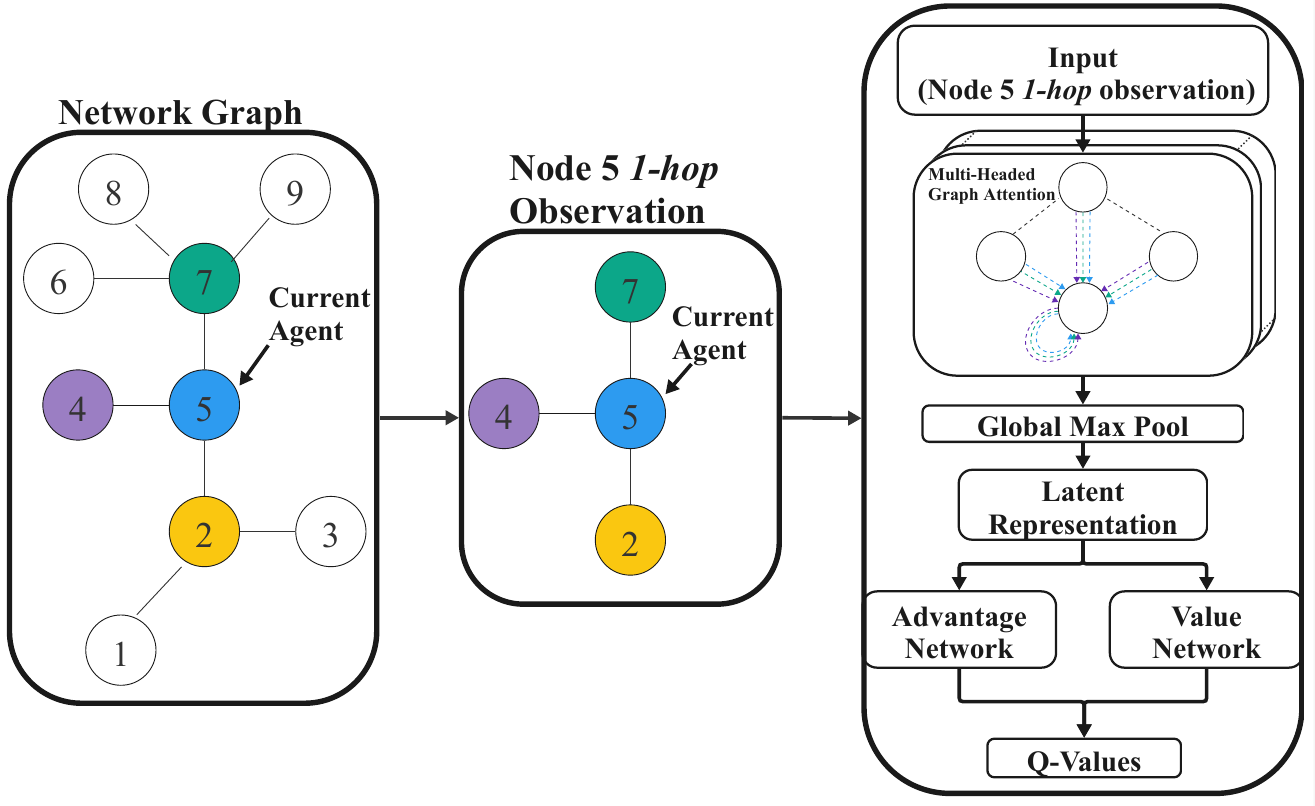}
    \caption{The \gls{hldyan} architecture.}
    \label{fig:learning_architecture-2}
\end{figure}

\subsubsection{Hyperlocal-DyAN} 
Intending to generate less communication overhead, we design a second model, named \gls{hldyan}, which resembles \gls{ldyan} in its form. We replace the three encoding stages with a single \gls{gat} layer with dynamic attention. Within agent $i$'s observation, we apply the \gls{gat} encoding process to every node, followed by a $\mathrm{ReLU}$ activation function. Finally, a global max-pooling layer is applied to summarize the most salient neighborhood characteristics, as shown in Figure \ref{fig:learning_architecture-2}.

The rationale for this approach is that agents can make informed decisions by processing their one-hop neighborhood dynamics from each neighbor's perspective, eliminating the need to share their latent representations, as seen in \gls{ldyan}.

In detail, agent $i$'s observation at time $t$ is fed to the \gls{gat} layer and, as opposed to \gls{ldyan}, such an operation is repeated for every node within the local observation of agent $i$, producing a set of latent representations comprising $\mathbf{\hat{Y}}_i$ and $\mathbf{\hat{Y}}_j, \forall j \in  \mathcal{N}_i(t)$. We then perform global max pooling, obtained through a feature-wise max operation:
\begin{equation}
\mathbf{H}_i = \mathrm{max}_{j \in \mathcal{N}_i(t) \cup \{ i \}} \, \mathbf{\hat{Y}}_j.
\end{equation}

Finally, $\mathbf{H}_i$ is fed to the dueling network following the same process described in Equation~\ref{eq:duel}.

%% file: sections/experimental_results.tex
\begin{table*}[t]
    \centering
    \input{tables/results_degree}
    \caption{Evaluation of \gls{ldyan}, \gls{hldyan}, \gls{mpr}, and \gls{dgn} in terms of Coverage and Messages forwarded involving different scenarios.}
    \label{tab:results}
\end{table*}

\paragraph{Experimental Setup}
We generated 50,000 connected graph topologies for training, each consisting of 50 nodes with a broadcasting range of 20 $\mathrm{units}$ and no constraints on the number of neighbors. For every learning algorithm, training was conducted five times adopting different random seeds (4, 9, 17, 42, 43) for 1 million agent steps. In each training episode, the environment randomly selected a graph as the initial graph and a node $n_s$ as the source. To mitigate strong mobility pattern biases, a random state generator determined the nodes' movements, which was seeded anew at the beginning of every episode. The nodes' speed is defined in terms of  $\mathrm{distance}$  $\mathrm{units}$ per $\mathrm{step}$ $(\frac{\mathrm{units}}{\mathrm{step}})$ and was set to 6 during training. 

Furthermore, we utilized 4 distinct sets of connected starting graph topologies, not seen during training, for testing purposes. These sets comprised 50 nodes per graph with various constraints on the maximum node degree allowed (5, 10, 25, and 49). Our evaluation process involved testing each graph 50 times and selecting a different node as the source $n_s$ in each iteration. To promote reproducibility and ensure the coherence of results, the same random state generator was used to control the nodes' movements across iterations for the same graph. Additional evaluations were conducted on the impact of the nodes' velocity setting their speed to $1$, $6$, and $10$ $\frac{\mathrm{units}}{\mathrm{step}}$.

Our analysis compared \gls{ldyan} and \gls{hldyan} with the \gls{mpr} heuristic and \gls{dgn}. The \gls{dgn} methodology excluded Temporal Relation Regularization, as it was unnecessary in our setting where agent interaction is temporally bounded by a short local horizon. To ensure a fair evaluation, we maintained consistent hyperparameters across all models, the details of which are presented in Table~\ref{tab:hyperparameters}-Appendix~\ref{subsubsec:hyperparameters}.



\begin{figure}[t]
  \centering
  \includegraphics[width=\columnwidth]{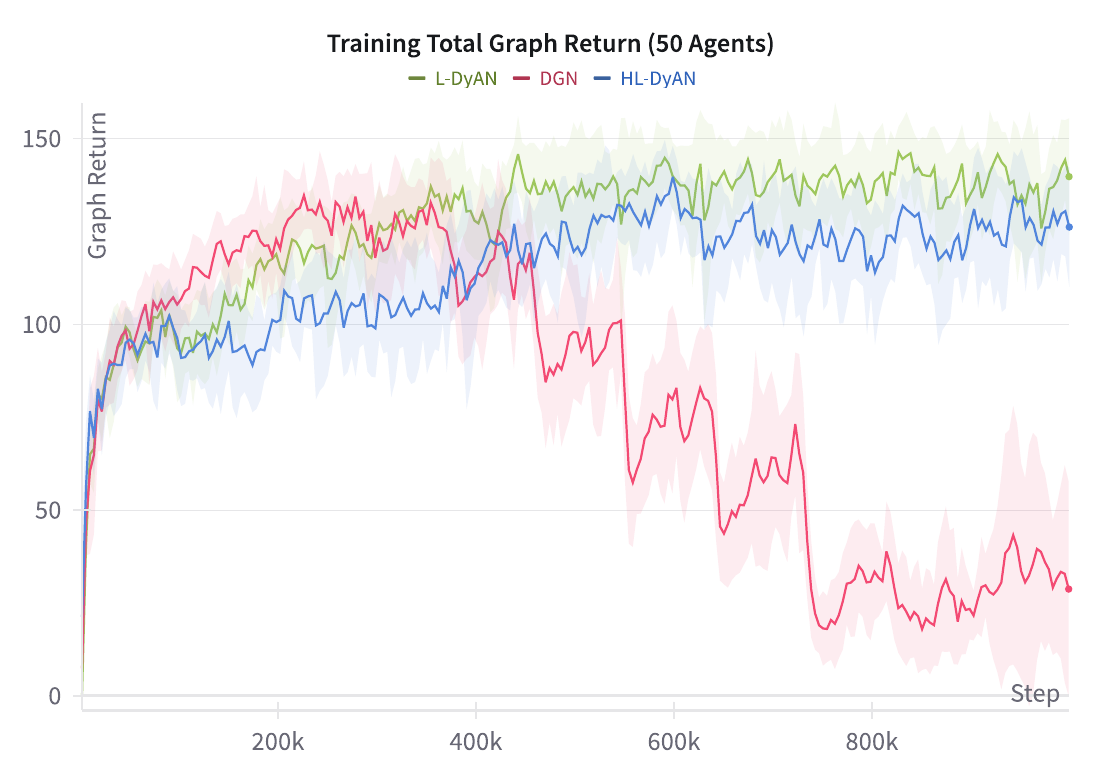}
  \caption{Sum of the returns during training with different seeds. Colored shades represent the respective standard deviations.}
  \label{fig:return}
\end{figure}

\paragraph{Results}
Table~\ref{tab:results} shows the results of our experiments in terms of coverage and messages required to achieve it, presenting the means and standard deviations.
Our proposed methods, \gls{ldyan} and \gls{hldyan}, consistently demonstrate higher coverage across different scenarios compared to \gls{mpr} and DGN. In particular, \gls{hldyan} achieves the highest coverage with a mean of 90.37\%. However, this comes at the cost of a higher number of messages, with \gls{hldyan} generating an average of 34.85 messages per episode. \gls{ldyan}, while slightly less successful in coverage (87.70\%), requires significantly fewer messages (24.32), indicating its suitability in scenarios where message efficiency is prioritized over coverage.

As the nodes' speed increases, the performance gap between our proposed methods and \gls{mpr} widens, with \gls{ldyan} and \gls{hldyan} maintaining superior coverage across all tests with increased speed. This is further supported by additional tests we conducted with the nodes' speed set to 10, obtaining a coverage of $82.35\%$, resp. $88.83\%$ with $23.88$ resp. $36.61$ messages, by \gls{ldyan} and \gls{hldyan}. \gls{mpr}, instead, struggled to reach $27\%$ of coverage.
Additionally, in the more dynamic scenarios with speed set to be greater than 1, the maximum node degree negatively influences the performance of all methods, but \gls{ldyan} and \gls{hldyan} consistently outperform the other algorithms, which fail to reach 50\% coverage.
This indicates the robustness of \gls{ldyan} and \gls{hldyan} to dynamic scenarios with less dense neighborhoods. 

In more static scenarios where the node speed is set to 1,  \gls{ldyan} and \gls{mpr} reveal to be the more efficient reaching, respectively, 14.84 and 14.54 percent of coverage per message.

Figure~\ref{fig:return} illustrates the training progress of \gls{ldyan}, \gls{hldyan}, and \gls{dgn} over multiple cycles using five distinct random seeds. Our proposed methodologies, \gls{ldyan} and \gls{hldyan}, demonstrated an average of the total graph returns (sum of all the agents returns) of $141.47 \pm 15.63$ and $127.50 \pm 13.57$, respectively. In contrast, the training trajectory of \gls{dgn} indicates a more brittle learning progress, unable to learn an effective multi-agent strategy for this task.

\paragraph{Discussion}
\label{subsec:discussion}

The results underscore the efficiency of \gls{ldyan} and \gls{hldyan} in learning effective multi-agent strategies balancing message efficiency with coverage consistently across various scenarios. \gls{mpr} falls short in both slightly and very dynamic and/or sparsely connected environments, with its performance worsening as nodes are faster and their starting neighborhood more sparse. The results also highlight the adaptability of \gls{ldyan} and \gls{hldyan} in varying network densities and node velocities, making them suitable for a wide range of dynamic network environments. Additionally, the training behavior of the three learning algorithms highlights the strengths of our method in learning multi-agent strategies for networked environments, where the presence of the agents and the structure of their neighborhoods change dynamically.

%% file: tables/results_degree.tex
\small
\begin{tabularx}{\textwidth}{l|c|c|c|c|c|c}
\toprule
Max Node Degree (Start) & Nodes Speed & Metric & L-DyAN & HL-DyAN & MPR Heuristic & DGN \\ \midrule

\multirow{2}{*}{5 Neighbors} &  \multirow{2}{*}{$1\frac{\mathrm{unit}}{\mathrm{step}}$} & Messages & 28.22 $\pm$ 4.52 & 34.81 $\pm$ 8.27 & 25.64 $\pm$ 7.47 & 3.88 $\pm$ 2.94 \\ 
& & Coverage &\textbf{93.78\% $\pm$ 13.58} & 90.02\% $\pm$ 19.93 & 86.24\% $\pm$ 22.40 & 24.34\% $\pm$ 16.24 \\ \hline

\multirow{2}{*}{5 Neighbors} &  \multirow{2}{*}{$6\frac{\mathrm{units}}{\mathrm{step}}$} & Messages & 24.05 $\pm$ 6.95 & 33.40 $\pm$ 8.23 & 8.21 $\pm$ 5.85 & 6.69 $\pm$ 4.39 \\ 
& & Coverage & 79.34\% $\pm$ 21.64 & \textbf{83.64\% $\pm$ 19.99} & 34.02\% $\pm$ 20.81 & 35.77\% $\pm$ 21.15 \\ \midrule

\multirow{2}{*}{10 Neighbors} & \multirow{2}{*}{$1\frac{\mathrm{unit}}{\mathrm{step}}$} & Messages & 23.26 $\pm$ 5.67 & 33.06 $\pm$ 7.43 & 22.95 $\pm$ 7.26 & 15.91 $\pm$ 2 \\
& & Coverage & 88.51\% $\pm$ 19.46 & \textbf{89.59\% $\pm$ 19.65} & 85.73\% $\pm$ 23.63 & 24.35\% $\pm$ 13.54 \\ \hline

\multirow{2}{*}{10 Neighbors} & \multirow{2}{*}{$6\frac{\mathrm{units}}{\mathrm{step}}$} & Messages & 24.31 $\pm$ 5.38 & 37.01 $\pm$ 5.26 & 8.10 $\pm$ 6.14 & 7.03 $\pm$ 4.57 \\
& & Coverage & 86.33\% $\pm$ 16.99 & \textbf{91.69\% $\pm$ 12.67} & 37.98\% $\pm$ 23.55 & 42.74\% $\pm$ 24.45 \\ \midrule

\multirow{2}{*}{25 Neighbors} & \multirow{2}{*}{$1\frac{\mathrm{unit}}{\mathrm{step}}$} & Messages & 23.74 $\pm$ 5.85 & 34.15 $\pm$ 4.72 & 24.93 $\pm$ 5.35 & 3.28 $\pm$ 2.23 \\
& & Coverage & 90.44\% $\pm$ 18.80 & \textbf{93.84\% $\pm$ 9.85}& 92.86\% $\pm$ 15.98 & 26.33 $\pm$ 16.02 \\ \hline

\multirow{2}{*}{25 Neighbors} & \multirow{2}{*}{$6\frac{\mathrm{units}}{\mathrm{step}}$} & Messages & 24.29 $\pm$ 5.67 & 36.35 $\pm$ 4.64 & 10.03 $\pm$ 6.75 & 6.96 $\pm$ 4.48 \\
& & Coverage & 88.19\% $\pm$ 17.62 & \textbf{92.23\% $\pm$ 11.06}& 46.80\% $\pm$ 27.05 & 44.02\% $\pm$ 25.50 \\ \midrule

\multirow{2}{*}{49 Neighbors} & \multirow{2}{*}{$1\frac{\mathrm{unit}}{\mathrm{step}}$} & Messages & 22.99 $\pm$ 5.85 & 34.34 $\pm$ 4.82 & 23.92 $\pm$ 6.97 & 3.47 $\pm$ 2.36 \\
& & Coverage & 89.73\% $\pm$ 20.02 & \textbf{91.96\% $\pm$ 18.04} & 88.93\% $\pm$ 20.04 & 27.16\% $\pm$ 15 \\ \hline

\multirow{2}{*}{49 Neighbors} & \multirow{2}{*}{$6\frac{\mathrm{units}}{\mathrm{step}}$} & Messages & 24.81 $\pm$ 4.95 & 36.61 $\pm$ 5.78 & 9.39 $\pm$ 6.57 & 6.81 $\pm$ 4.14 \\ 
& & Coverage & 88.86\% $\pm$ 15.34 & \textbf{92.42\% $\pm$ 13.51} & 43.49\% $\pm$ 25.67 & 41.33\% $\pm$ 22.61 \\ \bottomrule


\end{tabularx}

%% file: sections/Supplementary.tex
\clearpage

\section{Technical Appendix}
\bigskip
\subsection{Pseudo-code of the MPR Selection algorithm}
\label{sec:mpr-pseudo}
\begin{algorithm}[htb]
\caption{MPR Selection Heuristic}
\label{alg:mpr-selection}
    \begin{algorithmic}[1]
        \Require The set $N$ of one-hop neighbors
        \Ensure The MPR set
        \State Initialize MPR set with all members of $N$ with willingness equal to \texttt{WILL\_ALWAYS}
        \For{each node $y \in N$}
            \State Calculate $D(y)$
        \EndFor
        \State Select nodes in $N$ which cover the poorly covered nodes in $N_2$. Remove these nodes from $N_2$.
        \While{nodes exist in $N_2$ not covered by at least \texttt{MPR\_COVERAGE} nodes in the MPR set}
            \For{each node in $N$}
                \State Calculate reachability: number of nodes in $N_2$ not yet covered by at least \texttt{MPR\_COVERAGE} nodes in the MPR set and are reachable through this 1-hop neighbor.
            \EndFor
            \State Select as MPR the node with the highest willingness among nodes in $N$ with non-zero reachability.
            \If{multiple choices}
                \State Select node providing maximum reachability to nodes in $N_2$.
                \If{multiple nodes provide same reachability}
                    \State Select node as MPR with greater $D(y)$.
                \EndIf
            \EndIf
            \State Remove nodes from $N_2$ now covered by \texttt{MPR\_COVERAGE} nodes in the MPR set.
        \EndWhile
    \end{algorithmic}
\end{algorithm}
$D(y)$  is defined as the number of symmetric neighbors of node $y$, excluding all the members of $N$ and excluding the node performing the computation.
A poorly covered node is a node in $N_2$ which is covered by less than \verb|MPR_COVERAGE| nodes in $N$.
Note that in our implementation every node has willingness set to \verb|WILL_ALWAYS| and \verb|MPR_COVERAGE| is set to 1 to ensure that the MPR heuristic's overhead is kept to the minimum.

\begin{table*}[t]
    \centering
    \input{tables/results_extended}
    \caption{Evaluation of \gls{ldyan}, \gls{hldyan}, \gls{mpr}, and \gls{dgn} in terms of Coverage and Messages forwarded involving different scenarios.}
    \label{tab:results_extended}
\end{table*}
\subsection{Extended Results}
In Table~\ref{tab:results_extended} we present an extended version of Table~\ref{tab:results}, where we include additional networking scenarios with node velocities set to $3$ and $10 \frac{\mathrm{units}}{\mathrm{step}}$, for all the node degrees. We note that all of the additional experiments support the claims made in the paper.

\label{sec:ablation_study}
\input{sections/ablation_study}

\subsection{Additional Reproducibility Details and Instructions}
\label{sec:implementation}

\subsubsection{Implementation Details}
Our framework, which is written in Python and based on PyTorch, implements a customized extension of Tianshou~\cite{tianshou}. The \gls{marl} environment is defined following the PettingZoo~\cite{terry2021pettingzoo} API. The \gls{gat}, transformer-like dot product attention layer, and global max pooling follow the implementation provided by PyTorch Geometric~\cite{Fey/Lenssen/2019}. Training and testing graphs were generated with the aid of the NetworkX library~\cite{osti_960616}.

\subsubsection{Hardware Involved} 
Our policies were trained using 40 parallel environments on a workstation running Ubuntu 22.04 LTS, CUDA Toolkit v11.7, and equipped with an Intel i9-13900F CPU, 32GB DDR4 RAM, and an NVIDIA GeForce RTX 4090 GPU.

\subsubsection{Hyperparameters}
\label{subsubsec:hyperparameters}
\input{sections/training_performance}
    

\subsubsection{Instructions}

To ease testing and reproducibility, our framework comes with a Docker Image comprising all the requirements that the reader can quickly build, allowing them to run the application in a containerized environment following the instructions listed below. 
\begin{itemize}
    \item[1)] From the root project folder run the following command to build the image:
\begin{verbatim}
docker build -t marl_mpr .
\end{verbatim}
    \item[1b)] If the host machine does not have any GPU or if it is an Apple Mac device (including ones employing Apple MX SoC) please use:
\begin{verbatim}
docker build -t marl_mpr \ 
\end{verbatim}
\begin{verbatim}
    -f Dockefile.cpu .
\end{verbatim}
    \item[2)] Command example to run the container:
\begin{verbatim}
docker run --ipc=host --gpus all \
\end{verbatim}
\begin{verbatim}
    -v ${PWD}:/home/devuser/dev:Z \
\end{verbatim}
\begin{verbatim}
    -it --rm  marl_mpr
\end{verbatim}
Please note that \verb|--gpus all| is optional and it should be omitted in case the hosting machine is not equipped with a GPU.
Alternatively, the reader can also install the requirements needed in a dedicated Python ($\ge 3.8$) virtual environment by running \begin{verbatim}
pip install -r requirements.txt
\end{verbatim}
    \item[3)]\textbf{Visualization} \textit{(Optional)}. A visualization aid is provided to watch the agents in action on the testing graphs. If the Docker Image is used, the following two arguments should be added when running the container in order to render the figure: \verb|-e DISPLAY=unix\$DISPLAY|  and \verb|-v /tmp/.X11-unix:/tmp/.X11-unix|.\footnote{Please note that these arguments are valid only for machines running a Unix OS. Machines running MacOS might require installing a display server like XQuartz.}
    The entire command to run the container while enabling visualization would be:
    \begin{verbatim}
docker run --ipc=host --gpus all \
\end{verbatim}
\begin{verbatim}
    -e DISPLAY=unix$DISPLAY \
\end{verbatim}
\begin{verbatim}
    -v /tmp/.X11-unix:/tmp/.X11-unix \
\end{verbatim}
\begin{verbatim}
    -v ${PWD}:/home/devuser/dev:Z \
\end{verbatim}
\begin{verbatim}
    -it --rm  marl_mpr
\end{verbatim}
\end{itemize}

\subsubsection{Training models}
Trained models will be saved in the \\
\verb|log/algorithm_name/weights| folder as \verb|model_name_last.pth|. Before the training process begins, the user will be asked if they want to log training data using the Weight and Biases (WANDB) logging tool.

\begin{itemize}
    \item DGN
\begin{verbatim}
    python train_dgn.py \
\end{verbatim}
\begin{verbatim}
        --model-name DGN
\end{verbatim}
    
    \item L-DyAN
\begin{verbatim}
    python train_l_dyan.py \
\end{verbatim}
\begin{verbatim}
        --model-name L-DyAN \
\end{verbatim}
    
    \item HL-DyAN
\begin{verbatim}
    python train_hl_dyan.py \
\end{verbatim}
\begin{verbatim}
        --model-name HL-DyAN 
\end{verbatim}
\end{itemize}

\paragraph{Seeding} Results reported are based on 5 different random seeds $\left(4, 9, 17, 42, 43\right)$. By default, our scripts set the seed to \textit{9}. The reader can easily change such value using the argument \verb|--seed X|, where X is the chosen seed. This seeding value is carefully set for \verb|np_random.seed(X)|, \verb|toch.manual_seed(X)|, \verb|train_envs.seed(X)|, and \\ \verb|test_envs.seed(X)|. Other parameters can be changed from their default and they can be consulted via \verb|python train_dgn.py --help|.

\subsubsection{Testing models}
All three of our trained models, DGN, L-DyAN, and HL-DyAN, are found in their respective subfolders of the \verb|/log| folder and results can be reproduced with the following command: 
\begin{itemize}
    \item DGN
    \begin{verbatim}
    python train_dgn.py --watch \
    \end{verbatim}
    \begin{verbatim}
        --model-name DGN.pth
    \end{verbatim}
    \item L-DyAN
    \begin{verbatim}
    python train_l_dyan.py --watch \
    \end{verbatim}
    \begin{verbatim}
        --model-name L-DyAN.pth
    \end{verbatim}
    \item HL-DyAN
    \begin{verbatim}
    python train_hl_dyan.py --watch \
    \end{verbatim}
    \begin{verbatim}
        --model-name HL-DyAN.pth
    \end{verbatim}
    \item MPR Heuristic. In order to test MPR heuristic results add the boolean argument \verb|--mpr-policy|. For example:
    \begin{verbatim}
    python train_hl_dyan.py --watch \
    \end{verbatim}
    \begin{verbatim}
        --model-name HL-DyAN.pth \
        --mpr-policy
    \end{verbatim}
\end{itemize}

\subsubsection{Topologies Dataset} As mentioned in the \nameref{sec:results} Training and testing sets contain, respectively, 50K and 130 graphs. The sole dataset takes up to 140MB compressed in a ZIP file with maximum compression level, for such a reason, and for reviewing purposes, we upload (along with the code) a reduced training set of topologies with 10000 graphs. The testing set is left unchanged. All the topologies can be found in the \verb|/graph_topologies| folder.

%% file: tables/results_extended.tex
\small
\begin{tabularx}{\textwidth}{l|c|c|c|c|c|c}
\toprule
Max Node Degree (Start) & Nodes Speed & Metric & L-DyAN & HL-DyAN & MPR Heuristic & DGN \\ \midrule

\multirow{2}{*}{5 Neighbors} &  \multirow{2}{*}{$1\frac{\mathrm{unit}}{\mathrm{step}}$} & Messages & 28.22 $\pm$ 4.52 & 34.81 $\pm$ 8.27 & 25.64 $\pm$ 7.47 & 3.88 $\pm$ 2.94 \\ 
& & Coverage &\textbf{93.78\% $\pm$ 13.58} & 90.02\% $\pm$ 19.93 & 86.24\% $\pm$ 22.40 & 24.34\% $\pm$ 16.24 \\ \hline

\multirow{2}{*}{5 Neighbors} &  \multirow{2}{*}{$3\frac{\mathrm{units}}{\mathrm{step}}$} & Messages & 26.21 $\pm$ 5.85 & 34.55 $\pm$ 7.35 & 14.33 $\pm$ 8.42 & 5.21 $\pm$ 3.80 \\ 
& & Coverage &87.78\% $\pm$ 18.53 & \textbf{88.65\% $\pm$ 17.74} & 54.53\% $\pm$ 28.27 & 30.49\% $\pm$ 19.33 \\ \hline

\multirow{2}{*}{5 Neighbors} &  \multirow{2}{*}{$6\frac{\mathrm{units}}{\mathrm{step}}$} & Messages & 24.05 $\pm$ 6.95 & 33.40 $\pm$ 8.23 & 8.21 $\pm$ 5.85 & 6.69 $\pm$ 4.39 \\ 
& & Coverage & 79.34\% $\pm$ 21.64 & \textbf{83.64\% $\pm$ 19.99} & 34.02\% $\pm$ 20.81 & 35.77\% $\pm$ 21.15 \\ \hline

\multirow{2}{*}{5 Neighbors} &  \multirow{2}{*}{$10\frac{\mathrm{units}}{\mathrm{step}}$} & Messages & 22.10 $\pm$ 7.17 & 32.50 $\pm$ 7.97 & 4.49 $\pm$ 3.18 & 9.34 $\pm$ 5.02 \\ 
& & Coverage & 71.96\% $\pm$ 21.47 & \textbf{79.78\% $\pm$ 19.75} & 21.54\% $\pm$ 12.63 & 45.32\% $\pm$ 22.64 \\\midrule

\multirow{2}{*}{10 Neighbors} & \multirow{2}{*}{$1\frac{\mathrm{unit}}{\mathrm{step}}$} & Messages & 23.26 $\pm$ 5.67 & 33.06 $\pm$ 7.43 & 22.95 $\pm$ 7.26 & 15.91 $\pm$ 2 \\
& & Coverage & 88.51\% $\pm$ 19.46 & \textbf{89.59\% $\pm$ 19.65} & 85.73\% $\pm$ 23.63 & 24.35\% $\pm$ 13.54 \\ \hline

\multirow{2}{*}{10 Neighbors} & \multirow{2}{*}{$3\frac{\mathrm{units}}{\mathrm{step}}$} & Messages & 24.34 $\pm$ 5.88 & 35.97 $\pm$ 5.98 & 16.10 $\pm$ 8.58 & 4.07 $\pm$ 3.04 \\
& & Coverage & 88.73\% $\pm$ 18.83 & \textbf{93.30\% $\pm$ 14.39} & 65.4\% $\pm$ 30.46 & 29.36\% $\pm$ 19.06 \\ \hline

\multirow{2}{*}{10 Neighbors} & \multirow{2}{*}{$6\frac{\mathrm{units}}{\mathrm{step}}$} & Messages & 24.31 $\pm$ 5.38 & 37.01 $\pm$ 5.26 & 8.10 $\pm$ 6.14 & 7.03 $\pm$ 4.57 \\
& & Coverage & 86.33\% $\pm$ 16.99 & \textbf{91.69\% $\pm$ 12.67} & 37.98\% $\pm$ 23.55 & 42.74\% $\pm$ 24.45 \\ \hline

\multirow{2}{*}{10 Neighbors} &  \multirow{2}{*}{$10\frac{\mathrm{units}}{\mathrm{step}}$} & Messages & 24.66 $\pm$ 5.38 & 35.98 $\pm$ 6.11 & 5.05 $\pm$ 3.69 & 9.72 $\pm$ 4.91 \\ 
& & Coverage & 82.32\% $\pm$ 16.70 & \textbf{88.06\% $\pm$ 14.62} & 26.61\% $\pm$ 15.36 & 52.15\% $\pm$ 23.58 \\ \midrule

\multirow{2}{*}{25 Neighbors} & \multirow{2}{*}{$1\frac{\mathrm{unit}}{\mathrm{step}}$} & Messages & 23.74 $\pm$ 5.85 & 34.15 $\pm$ 4.72 & 24.93 $\pm$ 5.35 & 3.28 $\pm$ 2.23 \\
& & Coverage & 90.44\% $\pm$ 18.80 & \textbf{93.84\% $\pm$ 9.85}& 92.86\% $\pm$ 15.98 & 26.33 $\pm$ 16.02 \\ \hline

\multirow{2}{*}{25 Neighbors} & \multirow{2}{*}{$3\frac{\mathrm{unit}}{\mathrm{step}}$}  & Messages & 24.33 $\pm$ 5.11 & 36.05 $\pm$ 4.57 & 17.73 $\pm$ 7.59 & 4.25 $\pm$ 3.08 \\
& & Coverage & 90.81\% $\pm$ 15.26 & \textbf{93.49\% $\pm$ 10.38}&  72.92\% $\pm$ 27.48 & 31.04\% $\pm$ 19.16 \\ \hline

\multirow{2}{*}{25 Neighbors} & \multirow{2}{*}{$6\frac{\mathrm{units}}{\mathrm{step}}$} & Messages & 24.29 $\pm$ 5.67 & 36.35 $\pm$ 4.64 & 10.03 $\pm$ 6.75 & 6.96 $\pm$ 4.48 \\
& & Coverage & 88.19\% $\pm$ 17.62 & \textbf{92.23\% $\pm$ 11.06}& 46.80\% $\pm$ 27.05 & 44.02\% $\pm$ 25.50 \\ \hline

\multirow{2}{*}{25 Neighbors} &  \multirow{2}{*}{$10\frac{\mathrm{units}}{\mathrm{step}}$} & Messages & 24.46 $\pm$ 5.36 & 35.78 $\pm$ 5.07 & 5.75 $\pm$ 4.41 & 9.50 $\pm$ 4.94 \\ 
& & Coverage & 84.79\% $\pm$ 15.76 & \textbf{89.87\% $\pm$ 11.98} & 30.32\% $\pm$ 19.28 & 51.89\% $\pm$ 24.83 \\ \midrule

\multirow{2}{*}{49 Neighbors} & \multirow{2}{*}{$1\frac{\mathrm{unit}}{\mathrm{step}}$} & Messages & 22.99 $\pm$ 5.85 & 34.34 $\pm$ 4.82 & 23.92 $\pm$ 6.97 & 3.47 $\pm$ 2.36 \\
& & Coverage & 89.73\% $\pm$ 20.02 & \textbf{91.96\% $\pm$ 18.04} & 88.93\% $\pm$ 20.04 & 27.16\% $\pm$ 15 \\ \hline

\multirow{2}{*}{49 Neighbors} &  \multirow{2}{*}{$3\frac{\mathrm{units}}{\mathrm{step}}$} & Messages & 23.9 $\pm$ 5.43 & 35.8 $\pm$ 6.13  & 16.45 $\pm$ 8.02 & 4.76 $\pm$ 3.27 \\ 
& & Coverage & 90.05\% $\pm$ 18.72 & \textbf{93.02\% $\pm$ 15.04} & 67.91\% $\pm$ 28.87 & 33.38\% $\pm$ 19.25 \\ \hline

\multirow{2}{*}{49 Neighbors} & \multirow{2}{*}{$6\frac{\mathrm{units}}{\mathrm{step}}$} & Messages & 24.81 $\pm$ 4.95 & 36.61 $\pm$ 5.78 & 9.39 $\pm$ 6.57 & 6.81 $\pm$ 4.14 \\ 
& & Coverage & 88.86\% $\pm$ 15.34 & \textbf{92.42\% $\pm$ 13.51} & 43.49\% $\pm$ 25.67 & 41.33\% $\pm$ 22.61 \\ \hline

\multirow{2}{*}{49 Neighbors} & \multirow{2}{*}{$10\frac{\mathrm{units}}{\mathrm{step}}$} & Messages & 23.88 $\pm$ 5.52 & 35.55 $\pm$ 5.16 & 4.81 $\pm$ 3.63 & 9.81 $\pm$ 4.81 \\
& & Coverage & 82.35\% $\pm$ 17.32 & \textbf{88.83\% $\pm$ 12.24} & 26.68\% $\pm$ 16.69 & 52.69\% $\pm$ 23.77 \\ \bottomrule

\multicolumn{2}{c|}{Total Mean $\pm$ Std} & Messages & \textbf{24.35 $\pm$ 5.69} & \textbf{35.12 $\pm$ 6.23} & 13.62 $\pm$ 6.47 & 6.92 $\pm$ 3.88 \\
\multicolumn{2}{c|}{Total Mean $\pm$ Std} & Coverage & \textbf{86.50\% $\pm$ 18.01} & \textbf{90.02\% $\pm$ 15.46} & 55.12\% $\pm$ 22.99 & 37.02\% $\pm$ 20.71 \\ \midrule

\multicolumn{7}{l}{Every experiment is performed over 500 different starting graphs. Results report mean and standard deviation.} \\ \bottomrule
\end{tabularx}

%% file: sections/ablation_study.tex
\subsection{Ablation Study}\label{sec:ablation}
\begin{figure}[t]
    \centering
    \includegraphics[width=\columnwidth]{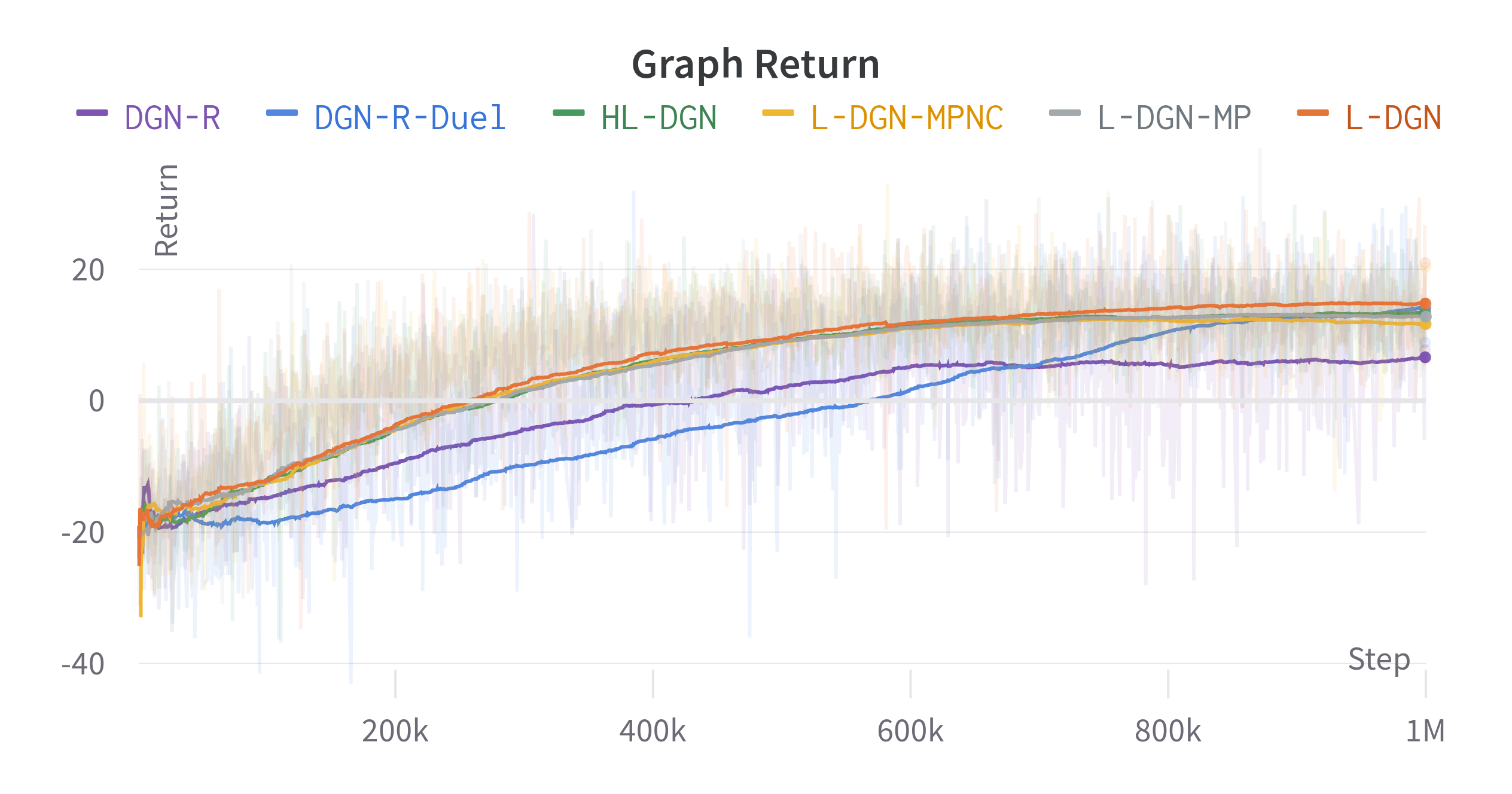}
    \caption{Graph return of the various methods used for the ablation study.}
    \label{fig:return_all}
\end{figure}

We investigate different ablations of \gls{ldyan}, whose architectures lay between \gls{ldyan} and \gls{hldyan}, and \gls{dgn}. 
Their performance is measured in terms of the summation of the returns achieved by each agent that has participated in the dissemination task, named \enquote{graph return} (Figure~\ref{fig:return_all}). Given that our environment is highly dynamic in terms of the entities contributing to the dissemination task at each timestep, such a metric allows us to understand if the local rewards assigned to each agent correlate with a desired overall collaboration across the entire graph, measured in terms of summations of the rewards achieved. We trained these policies on static scenarios with graphs of 20 nodes.

\textit{\gls{ldyan}-Duel.} The implementation of this method lies between \gls{ldyan} and DGN. Starting from the latter, we added the dueling network instead of a single \gls{mlp} stream as the action decoder. Figure~\ref{fig:return_all} shows the positive impact of the dueling network in the final strategy, which significantly outperforms \gls{ldyan} after 600K steps. From such a learning trajectory, we can also deduce the impact of another main component of our \gls{ldyan}, the n-step return estimation proportional to the local horizon (see \nameref{sec:local_horizon_estimation}). With the addition of such n-step returns, we obtain our \gls{ldyan} architecture, and we can notice how such a component helps the learned strategy to converge earlier and less abruptly.

\textit{\gls{ldyan}-MP.} This method removes the second \gls{gat} layer of \gls{ldyan} and replaces it with the global max pool operator (later adopted by \gls{hldyan}). The concatenation of the output of every encoding stage is still present here. We can notice a slight drop in performance when compared to \gls{ldyan}.

\textit{\gls{ldyan}-MPNC.} This method removes both the second \gls{gat} layer of \gls{ldyan}, as well as the concatenation of the output of every encoding stage. We notice a decrease in performance when compared to \gls{ldyan}. It can also be seen that \gls{hldyan} can be derived from \gls{ldyan}-MPNC after the ablation of the MLP encoding stage and that \gls{hldyan} does not suffer from such performance reduction.

In summary, these ablation studies centered around \gls{ldyan} allow us to both understand the strengths of this approach when compared to DGN, as well as motivate the design of the \gls{hldyan} architecture, which exhibits a simplified structure, less communication overhead, and only slightly underperforms in terms of graph return during training.  

%% file: sections/training_performance.tex
\begin{table}[h!]
    \centering
    \input{tables/hyperparameters}
    \caption{Hyperparameters used across our experiments. \enquote{Uniform} indicates that no prioritized replay has been employed.}
    \label{tab:hyperparameters}
\end{table}

%% file: tables/hyperparameters.tex
\begin{small}
\begin{tabularx}{\columnwidth}{X | c | c}
\toprule
    & Hyperparameter & Value \\
\midrule
    Training & & \\
    & Training steps & $1 \times 10^6$ \\
    & Learning rate & $1 \times 10^{-3}$ \\
    & Buffer size & $1 \times 10^{5}$ \\
    & Gamma & $0.99$ \\
    & Batch size & 32 \\
    & Exploration Decay & Exponential \\
    & Local Horizon & $4$ \\
    & N-Step Estimation & $4$ \\
    & Training Frequency & 1 per 160 Agent steps \\
    & Gradient Steps & $1$ \\
    & Parallel Training Envs & 40 \\
    & Experience Replay & Uniform \\
    & Seed & 4, 9, 17, 42, 43 \\
    
    Policy Params. & & \\
    & MLP Hidden Size & 32 \\
    & GAT Attention Heads & $4$\\
    & GAT Hidden Size & 32 (each head) \\
    & A-Network Hidden Sizes & [128, 128] \\
    & V-Network Hidden Sizes & [128, 128] \\
\bottomrule
\end{tabularx}
\end{small}